# Deep Neural Networks for ECG-free Cardiac Phase and End-Diastolic Frame Detection on Coronary Angiographies

Costin Ciusdel[1,2], Alexandru Turcea[1], Andrei Puiu[1,2], Lucian Itu[1,2], Lucian Calmac[3], Emma Weiss[4], Cornelia Margineanu[4], Elisabeta Badila[4], Martin Berger[5], Thomas Redel[5], Tiziano Passerini[6], Mehmet Gulsun[6], Puneet Sharma[6]

[1]Corporate Technology, Siemens SRL, B-dul Eroilor nr. 3A, 500007 Brasov, Romania
[2]Automation and Information Technology, Transilvania University of Brasov, Mihai Viteazu nr. 5, 5000174 Brasov, Romania
[3]Interventional Cardiology, Clinical Emergency Hospital, Calea Floreasca nr. 8, 014461 Bucharest, Romania
[4]Internal Medicine, Clinical Emergency Hospital, Calea Floreasca nr. 8, 014461 Bucharest, Romania
[5]Advanced Therapies, Siemens Healthcare GmbH, Siemensstr. 1, Bayern, 91301 Forchheim, Germany
[6]Digital Services, Digital Technology & Innovation, Siemens Healthineers, 755 College Road, Princeton, 08540 NJ, USA

**Abstract**
Invasive coronary angiography (ICA) is the gold standard in Coronary Artery Disease (CAD) imaging. Detection of the end-diastolic frame (EDF) and, in general, cardiac phase detection on each temporal frame of a coronary angiography acquisition is of significant importance for the anatomical and non-invasive functional assessment of CAD. This task is generally performed via manual frame selection or semi-automated selection based on simultaneously acquired ECG signals – thus introducing the requirement of simultaneous ECG recordings.
In this paper, we evaluate the performance of a purely image based workflow based on deep neural networks for fully automated cardiac phase and EDF detection on coronary angiographies. A first deep neural network (DNN), trained to detect coronary arteries, is employed to preselect a subset of frames in which coronary arteries are well visible. A second DNN predicts cardiac phase labels for each frame. Only in the training and evaluation phases for the second DNN, ECG signals are used to provide ground truth labels for each angiographic frame.
The networks were trained on 17800 coronary angiographies from 3900 patients and evaluated on 27900 coronary angiographies from 6250 patients. No exclusion criteria related to patient state (stable or acute CAD), previous interventions (PCI or CABG), or pathology were formulated. Cardiac phase detection had an accuracy of 97.6%, a sensitivity of 97.6% and a specificity of 97.5% on the evaluation set. EDF prediction had a precision of 97.4% and a recall of 96.9%. Several sub-group analyses were performed, indicating that the cardiac phase detection performance is largely independent from acquisition angles and the heart rate of the patient, and that the detection accuracy is slightly higher for the left coronary artery than for the right coronary artery. The average execution time of cardiac phase detection for one angiographic series was on average less than five seconds on a standard workstation. We conclude that the proposed image based workflow potentially obviates the need for manual frame selection and ECG acquisition, representing a relevant step towards automated CAD assessment.

## 1. Introduction

Invasive coronary angiography (ICA) is a diagnostic imaging procedure that provides important information about the structure and function of the heart, and represents the gold standard in Coronary Artery Disease (CAD) imaging [1]. It enables the assessment of the anatomical severity of coronary stenoses either visually or by computer-assisted quantitative coronary angiography (QCA) [2]. In view of the limitations of the pure anatomical evaluation of CAD, the functional index of Fractional Flow Reserve (FFR) has been introduced as an alternative [3], and recent technological advances also allow for image-based functional assessment of coronary stenoses based on ICA [4-6].
In this and other clinical settings based on the use of ICA, detection of EDF and, in general, cardiac phase detection on each temporal frame of a coronary angiography acquisition is of significant importance. Three-dimensional (3D) QCA is typically performed on end-diastolic angiographic frames (EDFs), when coronary artery motion is minimal [7]. The main input for image-based FFR computation is represented by a three-dimensional anatomical model of the coronary lumen of interest reconstructed from the segmentations performed on end-diastolic frames of two angiographic acquisitions at least 30° apart [3]. In the case of myocardial bridging, integral part of the diagnostic process is the identification of the systolic frames, in which the myocardium constricts a tunnelling coronary artery, potentially causing acute coronary syndrome [8]. Similarly, identification of diastolic frames is required for the accurate determination of maximal flow acceleration in aortocoronary artery (AC) bypass grafts, successfully assessed as a potential predictor of graft failure [9]. Finally, the determination of accurate diastolic flow velocities requires identification of the diastolic frames, and is instrumental to the assessment of diastolic coronary flow reserve, which was shown to be independently associated with the myocardial diastolic function [10].

Currently, the selection of the EDF and the identification of the cardiac phase are performed either manually or automatically based on simultaneously acquired ECG signals [5]. This has a number of drawbacks: ECG signals may not always be available, and cardiac phase detection based on ECG can be challenging if the signal-to-noise ratio is too low to accurately detect end-diastole or the signal presents artefacts [11], [12]. Methods for automated cardiac phase detection on medical images, without the need for processing ECG signals, have been described in the past for use cases involving cardiac echocardiographic images [13-17], cardiac angiographic images [18], cardiac MRI [19-20], and intravascular ultrasound [21]. Furthermore, a proof-of-concept study focusing on image based cardiac phase detection on invasive coronary angiography was previously proposed and evaluated on a dataset consisting of 15 patients [12].

Deep learning (DL), and specifically convolutional neural networks (CNNs), represent the state-of-the-art machine learning technique that has been shown to provide unprecedented performance at learning patterns in medical images, providing great promise for helping physicians during clinical decision making processes [22]. Previously reported deep learning related studies cover various types of problems, e.g. classification, detection, segmentation, for different types of structures, e.g. landmarks, lesion, organs, in diverse anatomical application areas [23].

In this paper, we introduce a methodology based on deep neural networks (DNN) for cardiac phase (systole or diastole) and EDF detection on X-ray coronary angiographic images. Ground truth labels, employed for training the model, are derived from simultaneously acquired ECG signals. The key features of our approach are: (i) a deep learning model performing both spatial and temporal convolutions for cardiac phase prediction, (ii) a deep learning model for identifying the angiographic frames on which coronary arteries are well visible, (iii) a comprehensive pre- and post-processing workflow for maximizing the performance, and generalizing the applicability of the methodology, and (iv) a very large database consisting of 45700 coronary angiographies images acquired from 10150 patients employed for training, validation and evaluation purposes.

## 2. Methods

Coronary arteries display significant motion on angiographic images during one heart cycle. The motion is the compound effect of cardiac contraction, respiratory motion, and possibly patient or table displacement, usually appearing as panning in angiographic images. Figure 1 illustrates the changing position of coronary artery anatomical landmarks during one heart cycle, while Figure 2 displays the motion of various landmarks during one heart cycle on several coronary angiographies. The main goal of the proposed workflow is to employ deep learning based techniques to determine the cardiac phase of each angiographic frame by implicitly analyzing the motion of arteries and structures visible on consecutive angiographic frames, and without using any associated ECG information.

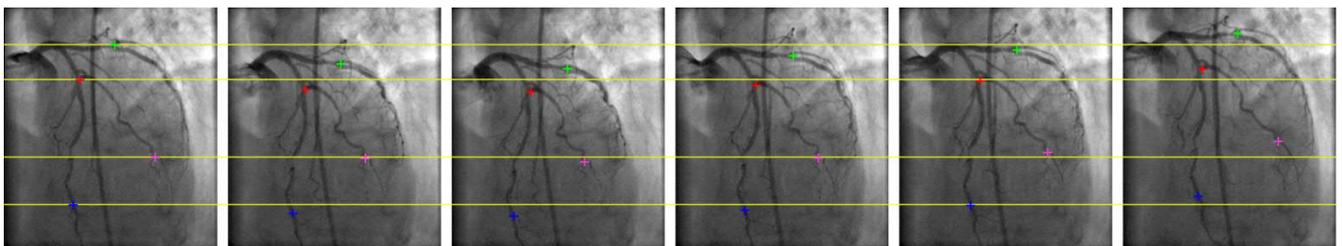

Figure 1: The changing position of coronary artery anatomical landmarks (bifurcations, sharp bends, stenoses, etc.) during one heart cycle.

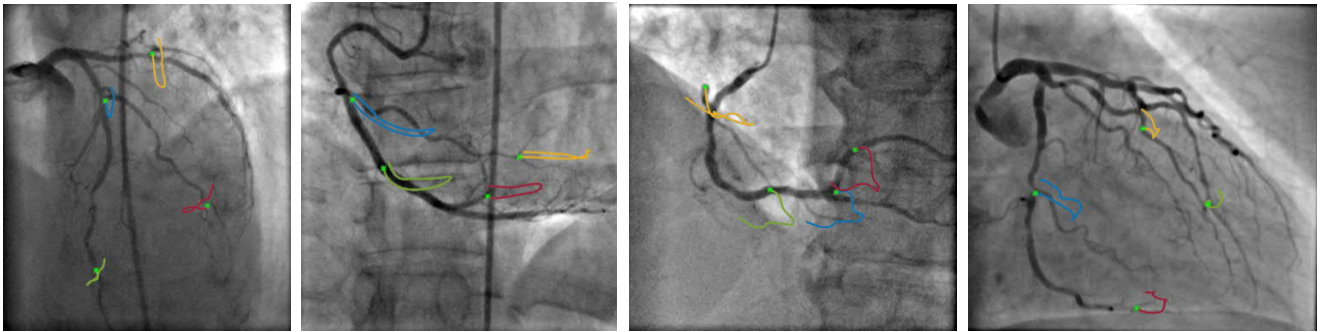
Figure 2: Motion patterns of various coronary artery anatomical landmarks (bifurcations, sharp bends, stenoses, etc.) during one heart cycle.

Figure 3 displays the overall workflow of the proposed methodology for cardiac phase detection, including offline training and online application. The individual steps depicted are described in detail in the sections below. In brief, for the online application, a first DNN, trained to detect coronary arteries, is employed to preselect a subset of frames in which coronary arteries are well visible. A second DNN predicts cardiac phase labels for each frame: it takes as input sequences of 10 frames from the preselected frame interval, performs spatial and temporal convolutions, and outputs predictions for the middle 4 frames of the sequence. The second DNN is applied with a sliding window mode approach. Only in the training and evaluation phases for the second DNN, ECG signals recorded simultaneously with angiographic images are used to provide ground truth labels: R-wave and T-wave peaks, employed for determining the onset and the end of systole respectively, are algorithmically detected, and the corresponding cardiac phase labels are mapped to each angiographic frame. The preliminary step of processing angiographic sequences to extract frames in which coronary vessels are clearly visible is also applied when building the training database for the deep learning method determining the cardiac phase of each angiographic frame.

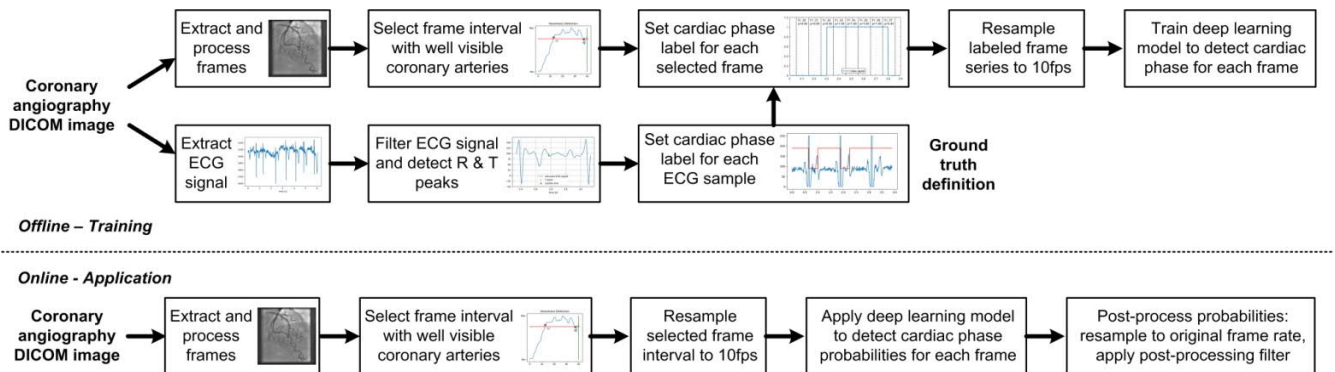
Figure 3: Overall workflow of the proposed methodology for offline training and online application.

**2.1. Offline training process**
**2.1.1. Pre-processing of ECG signals and angiographic frames**
Anonymized DICOM-formatted coronary angiographic acquisitions are used as input data. In the pre-processing stage, each DICOM file is parsed to extract the imaging data, i.e. angiographic frames are resampled to 512x512 pixel resolution if their original resolution is higher, and the raw ECG signal used for ground truth definition (Figure 4 left). Diastole-systole and systole-diastole transitions are detected to define the ground truth cardiac phase labels. For the former, BioSPPy, an open-source toolbox for biological signal processing, is employed [24]. First, BioSPPy applies a default processing pipeline to the original ECG signal, based on a dual-pass zero phase delay band pass linear finite impulse response filter. Next, the R peaks are detected as surrogate for diastole-systole transition (Figure 4 middle). The prerequisite for identifying the systole-diastole transition is the detection of the T peak. High frequency oscillations are typically still present after the first filtering step, hindering a precise T peak detection. Therefore, another stage of zero phase delay low pass filtering is applied. Finally, the T peaks are identified according to the following rules (Figure 4 right):

- The T peak time point should reside between two predefined limits, expressed relative to the heartbeat duration: herein, the limits of 20% and 65% were selected [25]
- The T peak should be a local maximum or minimum of the filtered ECG signal
- The T peak should be the local maximum or minimum with the largest temporal span, with respect to neighbouring maxima locations, in the considered window

Finally, the time point corresponding to the systole-diastole transition is defined as the first local minimum or maximum after the T peak, or as the time point where the filtered signal decreases or increases below the 20%-65% window mean value, whichever time point is encountered first [25]. A binary classification signal is then generated based on the detected time points: '0' for systole and '1' for diastole.

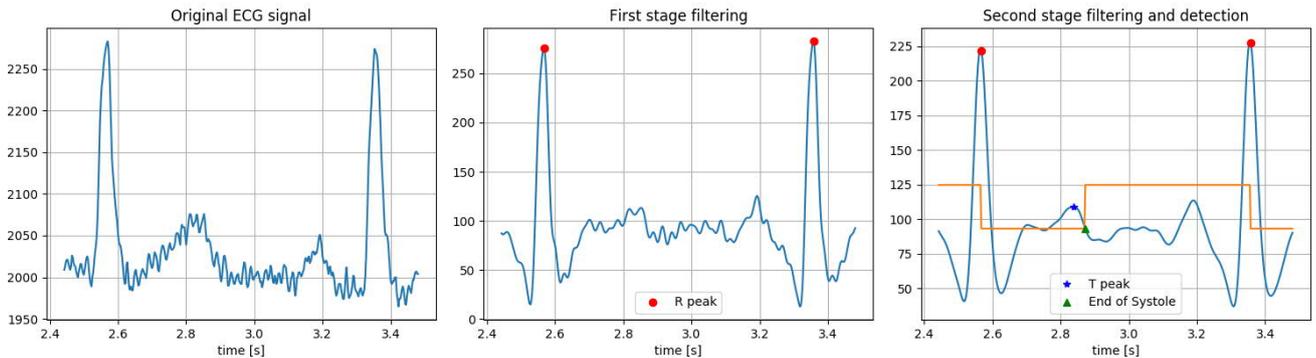

Figure 4: Processing of ECG signals: original ECG signal, first stage filtering, i.e. dual-pass zero phase delay band pass linear finite impulse response filter and R peak detection, second stage filtering, i.e. zero phase delay low pass filtering, and T peak and end of systole detection.

The angiographic frames extracted from the DICOM files are cropped and rescaled to remove the collimated areas of the frames, containing no relevant anatomical information. Typically, the angiographic image acquisition starts before injecting the contrast agent, and ends after the contrast agent has started to be washed out by the coronary arterial blood flow. The goal of the herein proposed method is to correctly predict the cardiac phase labels and detect EDFs for the angiographic frames on which coronary arteries are well visible, i.e. filled with contrast agent, as this corresponds to the set of clinically relevant frames for the applications that we are considering. For each frame the visibility of the coronary arteries is determined using a deep learning classifier trained to detect pixels which represent coronary arteries, i.e. to estimate the 'vesselness' of the image. All frames with visibility higher than a given threshold, defined relatively to the maximal number of pixels per frame representing coronary arteries, are marked as candidates; the largest continuous sequence of candidate frames is selected for further processing (see the appendix for more details).

Each angiographic frame in the selected interval is labelled using ECG-based labels defined as described above. Since angiographic sequences have lower temporal resolution compared to ECG signals (the typical sampling rate for ECGs stored in angiographic DICOMs is 400Hz, whereas the frame rate of angiograms is between 7.5 - 30 fps), we first subdivide the binary classification signal into N time intervals, where N is the number of frames in the coronary angiography. Then for each time interval we uniquely define the cardiac phase label for the corresponding angiographic frame using a voting scheme based on averaging. Intermediate values, between 0 and 1, are not rounded.

Furthermore, to develop a cardiac phase detection model which is independent from the acquisition frame rate, the labelled frame series is resampled at 10 fps. This was the minimal frame rate in the available datasets, if the frame rate is higher a subset of the frames is retained.

### 2.1.2. Deep learning model architecture and training process

The cardiac phase predictor is a deep neural network taking as input a sequence of angiographic frames (Figure 5). The output of the network is the classification of the middle frames of the input sequence, in terms of the probability of each frame to being a diastolic or systolic frame. Since we consider a binary classification problem, with the diastolic and systolic classes being associated with class index 1 and 0 respectively, the output probability can be interpreted as the predicted class index. We designed the network to process a sequence of 10 frames, covering exactly one second of acquisition. This choice was made as a trade off between accuracy of detection and memory usage during the training phase. The network outputs the classification only for the middle 4 frames of the sequence, so that for each frame the classification task uses information both from prior as well as following frames.

A shown in Figure 5, a first convolutional neural network (CNN) is employed to perform spatial convolutions, mapping each 512 x 512 input image to a 64 dimensional feature vector. The spatial CNN employs a classic structure of 2D convolutional layers followed by max pooling layers, and, finally, by a fully connected layer. The same CNN is applied independently to each frame, yielding a 10 x 64 feature matrix. A second CNN performs temporal convolutions on this matrix, independently for all 64 image features. Its output is an 8 x 64 matrix, which contains 8 temporal features for each one of the 64 spatial features. This matrix is then fed to a two layer classifier, which computes the output probabilities for the middle 4 frames in the input sequence (frames 4-7).

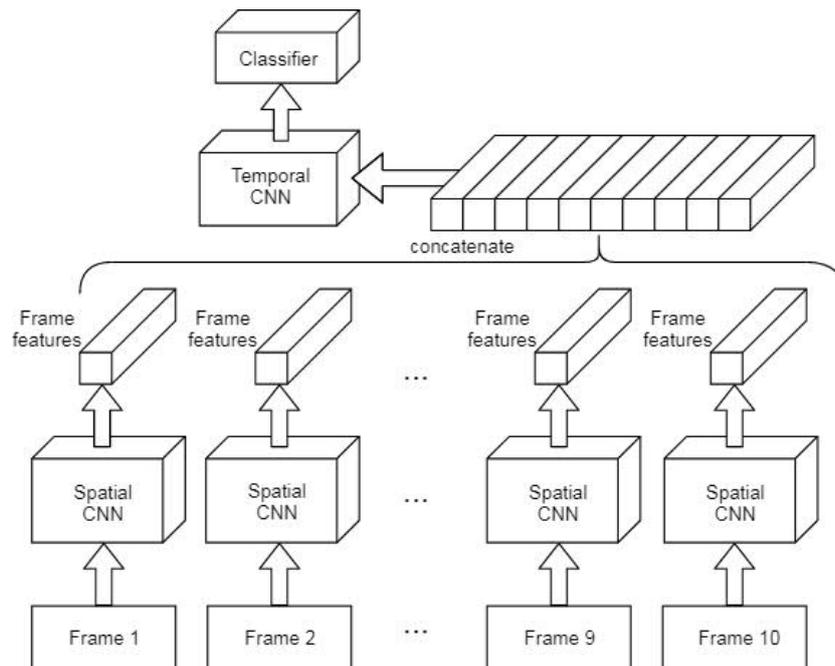

Figure 5: Overview of the deep neural network employed for cardiac phase detection.

After performing the selection of relevant frames using the vesselness model, and the resampling at 10 fps, the training datasets are generated as sequences of 10 frames. Considering $x$ the number of available frames:
- $x < 9$: no training dataset is generated
- $x \geq 10$: $x - 9$ training datasets are generated by considering all possible combinations of 10 consecutive frames

Ground truth labels for the middle 4 frames of each sequence are obtained by processing the associated ECG signal as described in the previous section.

Training is performed using the Adam [26] optimizer and a custom loss function based on the Poisson distribution divergence [27] which penalizes the output probability less significantly if the class is predicted correctly with respect to the 0.5 threshold value. Finally, we note that data augmentation techniques, as well as several dropout layers are used to increase the generalization capabilities of the model.

**2.2. Online application process**

During the online application of the trained deep learning model for cardiac phase prediction, no ECG information is used. The angiographic frames are pre-processed with the same methods used to populate the training database: they are cropped and rescaled, and the DL based vesselness estimator is employed to select the relevant frame interval. Next, the frames within the chosen interval are resampled at 10 fps and provided as input to the cardiac phase predictor.

To minimize the amount of redundant operations, first the spatial CNN is applied to all input frames, and the computed frame features are stored for further use. Next, the temporal CNN and the output classifier are applied with a sliding window approach, with step size one, using the stored frame features as input. Since the model outputs probabilities for the middle four frames, several frames end up with multiple predictions, and we consider for the final cardiac phase predictions only the frames for which at least two probabilities were generated. The final prediction is selected as the probability value closest to one of the class indices. In an additional post-processing step, the resulting sequence of probability values is resampled to the original frame rate using linear interpolation.

In case of panning, the table movement and the motion of the arteries and of other anatomical structures are superimposed, which lowers the confidence of the predictions, and can lead to spurious transitions between the two cardiac phases. To avoid such prediction artefacts a Schmitt trigger filter with adjustable thresholds is applied as a final post-processing step. The output is a binary classification signal, where zero corresponds to systole and one to diastole.

Figure 6 displays for a coronary angiography with a frame rate of 15fps the main steps performed during the online prediction phase. The angiographic frames are pre-processed, and the DL based vesselness estimator is employed to select the relevant frame interval (Figure 6A). Next, the selected frame interval is resampled at 10 fps, and the DL model for cardiac phase detection is applied with a sliding window approach, generating multiple prediction probabilities for each frame. The number of predictions depends on how many times the considered frame is included in one of the sliding windows. To increase the robustness of the classification, we only retain frames for which at least two predictions are generated, and for each frame only the probability value closest to one of the class indices is retained. In Figure 6B the up to 4 predictions are displayed for each frame alongside the finally selected probability. Finally, this sequence of probability values is resampled to the original frame rate and passed through a Schmitt trigger filter to determine the binary classification signal (Figure 6C). The ground truth labels are displayed in the upper left corner of each frame, the predicted label alongside the final probability are displayed in the lower left corner of each frame, and EDF frames are specifically marked. Figure 6 also displays a plot of the predicted probabilities and cardiac phase labels obtained after applying the post-processing filter, and a comparison of the predicted cardiac phase labels with the ECG based ground truth values.

## 3. Results

The workflow for cardiac phase detection was trained and evaluated using independent datasets. The training dataset consisted of 17800 coronary angiographies acquired from 3900 patients. An 80 – 20 % split was applied at patient level during training for setting up the actual training and validation datasets.

The main evaluation dataset consisted of 27900 coronary angiographies from 6250 patients, and was used for quantitatively assessing the performance of the cardiac phase and end-diastolic frame detection workflow, and for performing various subgroup analyses based on acquisition angles, heart rate, and vessel of interest. Furthermore, the ECG signals of a subset of coronary angiographies extracted from the main evaluation dataset were annotated by experienced clinical examiners, and the cardiac phase detection performance was assessed separately on this subset.

When setting up all datasets we considered all angiographic acquisitions of all patients which fulfilled the following inclusion criteria:
- Total number of frames > 20
- Number of frames with well visible arteries > 15
- ECG signal is present in the DICOM file
- At least two R peaks detected by the BioSPPy toolbox

In the training and the main evaluation dataset no exclusion criteria were formulated related to acquisition angles, heart rate, contrast agent intensity, patient characteristics, and patient diagnosis.

A secondary evaluation dataset, consisting of 175 coronary angiographies, with stricter inclusion and exclusion criteria was also considered, for assessing the performance as a function of the additional criteria. Finally, execution time considerations are briefly presented. The deep learning models were trained using Keras [28] with a TensorFlow [29] backend, while all pre- and post-processing steps were implemented in Python.

**Frames are pre-processed & DL based vesselness detector is employed to select the relevant frame interval**

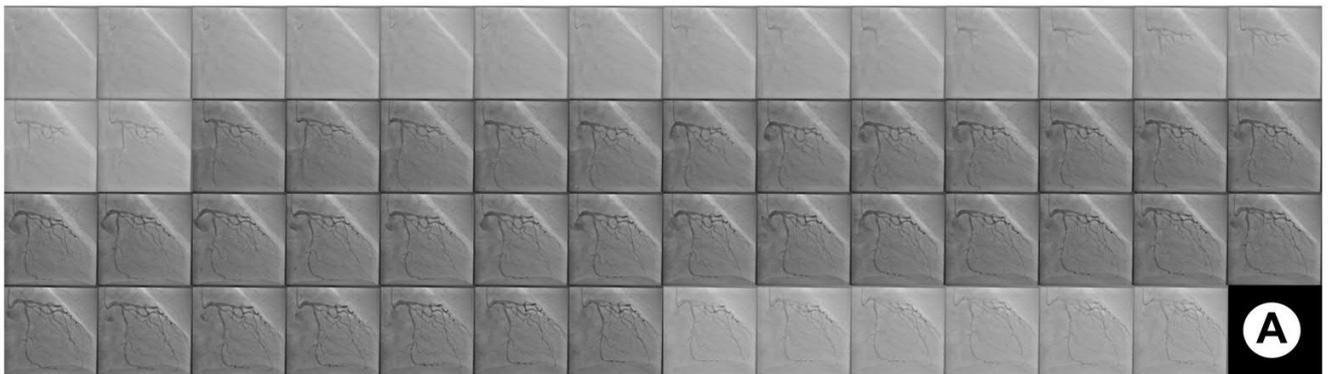

**Selected frame interval is resampled to 10fps and the DL model for cardiac phase detection is applied in a sliding windows mode approach**

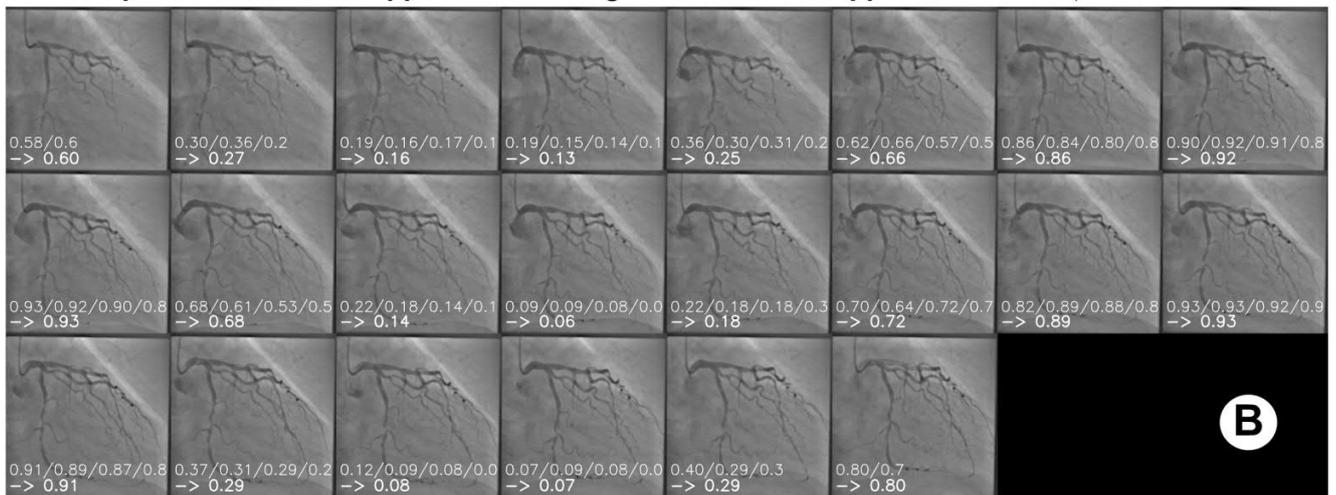

**Post-processing: resample to original frame rate, apply post-processing filter**

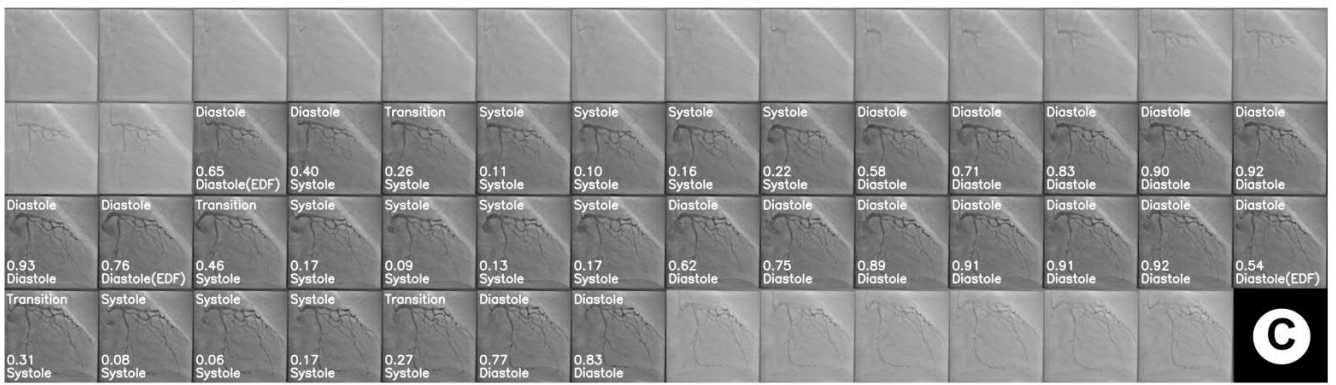

**Applying the post-processing filter on the upsampled probabilities**

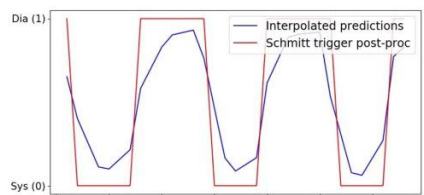

**Comparison of predicted cardiac phase labels with ECG based ground truth labels**

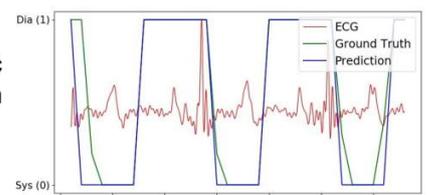

Figure 6: Main steps performed during the online prediction phase for a coronary angiography recorded at 15 fps. (A) The angiographic frames are pre-processed, and the DL based vesselness estimator is employed to select the relevant frame interval; (B) The selected frame interval is resampled at 10 fps and the DL model for cardiac phase

prediction is applied with a sliding window mode approach, generating multiple prediction probabilities for each frame, out of which only the probability closest to one of the class indices is retained; the up to 4 predictions are displayed for each frame alongside the finally selected probability, for the frames which have at least 2 predictions; (C) The predictions are resampled to the original frame rate and passed through a Schmitt trigger filter to determine the binary classification signal. The ground truth labels are displayed in the upper left corner of each frame, the predicted label alongside the final probability are displayed in the lower left corner of each frame, and EDF frames are specifically marked. A plot of the predicted probabilities and cardiac phase labels obtained after applying the post-processing filter, and a comparison of the predicted cardiac phase labels with the ECG based ground truth values are displayed.

**3.1. Cardiac phase detection performance on the main evaluation dataset**

Figure 7 displays sample comparisons of image based cardiac phase predictions and ECG based cardiac phase ground truth labels. While the image based cardiac phase classification performs well in all cases, a temporal offset can be observed at some transitions. The image-based classification is not available during the first and the last part of the acquisition, where the coronary arteries are not well visible. Noise and wave deformations are present in the ECG signals and can significantly influence the performance of the R and T peak detection algorithms.

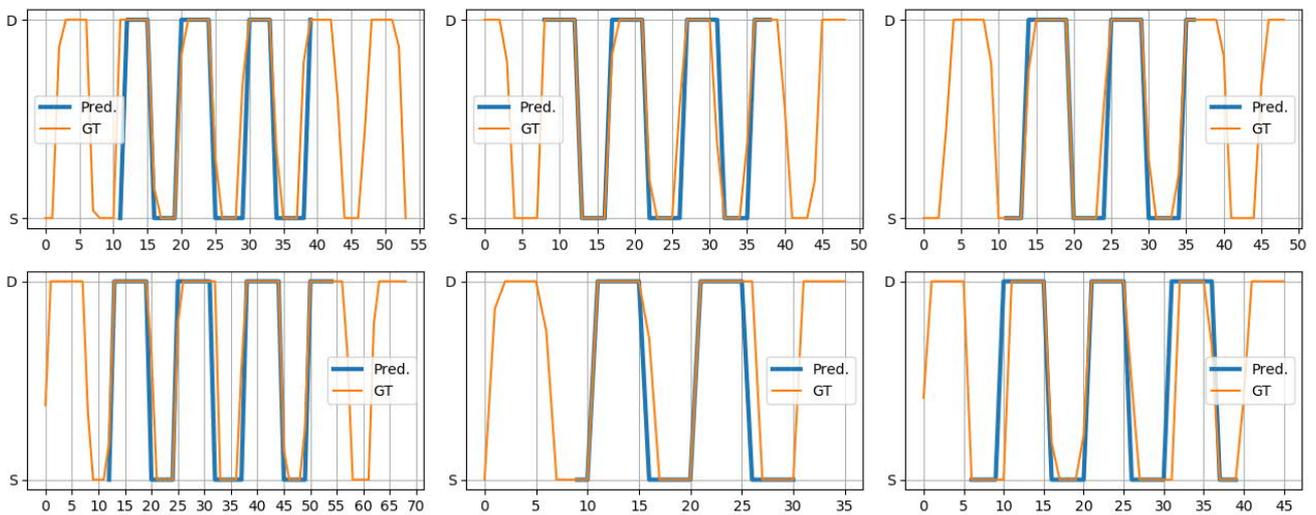

Figure 7: Sample comparisons of image based cardiac phase classifications and ECG based cardiac phase ground truth labels. 'S' stands for systole, 'D' for diastole, and X axis reports frame number).

For the quantitative evaluation of the cardiac phase detection workflow performance, we excluded frames with intermediate ground truth label values and frames adjacent to a cardiac phase transition.

Table 1 displays the statistical measures for cardiac phase detection on the evaluation dataset, under two settings: (a) all coronary angiographies have the same weight, i.e. statistics are first computed independently for each coronary angiography, and then an average value is computed for all angiographies, and (b) all frames have the same weight. Under the second setting, coronary angiographies with a greater number of frames have a larger weight than those with few frames, while under the first setting frames pertaining to coronary angiographies with fewer frames have a larger weight than frames pertaining to coronary angiographies with more frames. Statistical measures are slightly better under the second setting since coronary angiographies with few frames, on which the appearance of coronary arteries may be suboptimal, have a smaller weight than those with a large number of frames.

To assess the performance of the image based detection independently for the two cardiac phase transitions, systole → diastole and diastole → systole, we determined the cardiac phase misclassifications at frame level around each type of transition for the coronary angiographies with annotated ECG signals. We found that 65.0% / 65.2% of the errors corresponded to the systole → diastole transition, and 35.0% / 34.8% of the errors corresponded to the diastole → systole transition. This indicates that the diastole → systole transition is detected more reliably by the proposed workflow, possibly caused by:

- Onset of systolic contraction can be more precisely identified in the angiographic frames than the onset of diastolic relaxation
- Onset of systolic contraction is more precisely identified by the R peak, than the end of systole time point which is identified after the T peak

In the following we present various subgroup analyses, where the statistics are computed in the variant of all coronary angiographies having the same weight.

Table 1: Statistical measures on the evaluation dataset, computed in two variants: (a) all coronary angiographies have the same weight, and (b) all frames have the same weight.

| Statistical measure | Identical weights at coronary angiography level | Identical weights at frame level |
|---|---|---|
| Accuracy | 97.6% | 98.2% |
| Sensitivity | 97.6% | 98.2% |
| Specificity | 97.5% | 98.1% |
| PPV | 98.6% | 99.1% |
| NPV | 95.7% | 96.3% |

### 3.1.1. Cardiac phase detection performance vs. acquisition angles

Each coronary angiography is characterized by a primary (RAO - LAO) and a secondary (CAU - CRA) acquisition angle, which define the position of the X-Ray beam about the patient. Standard angiographic views, defined by acquisition angle ranges, have been developed for the visualization of the coronary arteries and their specific segments, to optimize lesion assessment, quantitative evaluation and stent selection [30].

Table 2 displays the cardiac phase detection performance for the generally accepted standard angiographic views, alongside the percentage of angiographies pertaining to each view. The detection accuracy is significantly lower only for the view enabling the left main ostium visualization.

Table 2: Cardiac phase detection performance for standard angiographic views on the evaluation dataset.

| Acquisition angles | Accuracy | % Angiographies | Target |
|---|---|---|---|
| LAO 40° – 60° CAU 10 – 30 | 95.2% | 6.0% | Left main, proximal LAD, proximal LCx |
| RAO 10 – 20 CAU 15 – 20 | 99.3% | 0.7% | Left main bifurcation, proximal LAD, proximal to mid LCx |
| RAO 0 – 10 CRA 25 – 40 | 98.7% | 13.3% | Mid and distal LAD, distal LCx, separate septals from diagonals |
| LAO 30 – 60 CRA 15 – 30 | 96.8% | 5.0% | Mid and distal LAD (left dominance), distal LCx, separate septals from diagonals |
| RAO 5 – LAO 5 CAU 5 – CRA 5 | 93.9% | 0.8% | Left main ostium |
| RAO 5 – LAO 5 CAU 20 – 30 | 98.7% | 8.3% | Distal left main bifurcation, proximal LAD, proximal to mid LCx |
| RAO 5 – LAO 5 CRA 25 – 35 | 98.9% | 12.2% | Proximal and mid LAD |
| LAO 20 – 40 CRA 15 – 25 | 97.4% | 4.4% | Proximal and mid RCA |
| RAO 20 – 40 CRA 15 – 25 | 99.0% | 0.7% | Proximal and mid RCA |
| RAO 5 – LAO 5 CRA 30 – 40 | 99.0% | 10.8% | Distal RCA |
| LAO 40 – 60 CRA 25 – 35 | 95.6% | 0.7% | Distal RCA |

### 3.1.2. Cardiac phase detection performance vs. heart rate

Given that the number of frames pertaining to the diastolic and systolic cardiac phase depends not only on the frame rate but also on the heart rate of the patient, we also analyzed the dependence of the cardiac phase detection performance on the patient's heart rate. An average heart rate value was computed for each coronary angiography based on the R peaks detected on the associated ECG signal: Table 3 displays the bin-based accuracies. The performance is stable across bins, with slightly lower accuracy at the two extremities. Naturally, the performance worsens for increased heart rates since the number of frames per individual cardiac phase drops and, consequently, also the number of frames which can be classified with high confidence decreases.

Table 3: Cardiac phase detection performance for different heart rate ranges

| Heart rate [bpm] | Accuracy | % Angiographies |
|---|---|---|
| 30 – 55 | 97.2% | 1.5 |
| 55-65 | 98.3% | 23.6 |
| 65-75 | 98.0% | 34.8 |
| 75-85 | 97.5% | 22.1 |
| 85-95 | 97.3% | 10.8 |
| > 95 | 93.5% | 7.2 |

### 3.1.3. Cardiac phase detection performance vs. vessel of interest

Coronary angiographies are recorded either for the left coronary artery (LCA) or the right coronary artery (RCA). Using a previously developed DL based classifier (98.8% accurate) we determined the coronary artery of interest for each angiography and computed the mean accuracy for each class: 98.4% for LCA, i.e. 73.3%% of all coronary angiographies, and 95.2% for RCA, i.e. 26.7% of all coronary angiographies. A similar LCA / RCA distribution could be observed in the training dataset. This indicates that the cardiac phase can be more reliably detected for LCA coronary angiographies, possibly caused by the fact that LCA angiographies contain a larger number of vessels.

### 3.1.4. Cardiac phase detection performance vs. expert annotations

To further validate our approach, we considered a subset of 300 patients randomly chosen from the main evaluation dataset. For each of the patients we randomly picked one angiographic acquisition which fulfilled the inclusion criteria, and two experienced clinical examiners independently annotated all R and T peaks on the corresponding ECG signals. Since T peaks could not always be reliably detected on the ECG signals extracted from the angiographic images, we finally only retained those angiographic acquisitions on which R and T peaks were identified by both clinical examiners, leading to an evaluation dataset consisting of 207 coronary angiographies. Starting from the R and T peak annotations we defined the ground truth cardiac phase labels as described in the methods section. Table 4 displays the statistical measures for cardiac phase detection on this evaluation dataset. Overall, the statistical measures obtained for this small evaluation dataset are comparable to those obtained on the entire evaluation dataset on which R and T peaks were determined algorithmically. No significant inter-observer variability was noted between the two clinical examiners.

Table 4: Statistical measures on the evaluation dataset consisting of coronary angiographies for which the ECG signals were annotated by two experienced clinical examiners.

| Statistical measure | GT defined by clinical examiner 1 | GT defined by clinical examiner 2 |
|---|---|---|
| Accuracy | 97.6% | 97.6% |
| Sensitivity | 97.2% | 97.3% |
| Specificity | 98.3% | 98.3% |
| PPV | 99.3% | 99.3% |
| NPV | 93.6% | 93.5% |

### 3.1.5. End-diastolic frame detection performance on the main evaluation dataset

As described in the introduction, one of the goals for performing cardiac phase detection on coronary angiographies is the detection of EDFs suitable for 3D QCA and coronary artery segmentations. Table 5 displays the statistical measures for EDF detection on the main evaluation dataset, obtained with a tolerance range of ±1 frames around the ground-truth EDF as detected based on the ECG signal. For the image based predictions, each diastolic frame which is followed by a systolic frame is considered to be an EDF. For the ground truth signal, each diastolic frame which is followed by a transition frame, or a systolic frame is considered to be an EDF.

Table 5: Statistical measures for EDF detection, with a tolerance range of ±1 frames around the ground-truth EDFs as detected based on the ECG signal.

| Statistical measure | Performance |
|---|---|
| Precision | 97.4% |
| Recall | 96.9% |
| F1 score | 97.2% |

### 3.2. Cardiac phase detection performance on the secondary evaluation dataset

To further validate our approach, we also considered a secondary evaluation dataset containing 175 coronary angiographies acquired from 84 patients, which were used in a recently published study focusing on computing FFR with a machine learning based approach from rest-state pressure measurements and anatomical features of the arteries and stenoses of interest extracted from the angiographic images [31]. These coronary angiographies had additional inclusion and exclusion criteria, related to:
- Coronary angiography acquisition: no panning, coronary arteries well visible along multiple heart cycles, no significant vessel segment overlapping and foreshortening
- Patient characteristics: stable CAD, no chronic total occlusions, no previous coronary artery bypass graft, no significant arrhythmia

Ground truth cardiac phase classification at angiographic frame level was performed algorithmically as described in the methods section. Table 6 displays the statistical measures for cardiac phase detection on this evaluation dataset, under two settings: (a) all angiographies have the same weight, and (b) all frames have the same weight. When compared to the results in Table 1, we note that the statistical measures are superior.

Table 6: Statistical measures on the secondary evaluation dataset, computed in two variants: (a) all coronary angiographies have the same weight, and (b) all frames have the same weight.

| Statistical measure | Identical weights at coronary angiography level | Identical weights at frame level |
|---|---|---|
| Accuracy | 98.5% | 98.9% |
| Sensitivity | 98.6% | 99.5% |
| Specificity | 98.3% | 98.1% |
| PPV | 98.6% | 98.8% |
| NPV | 98.4% | 99.2% |

Finally, Table 7 displays the statistical measures for EDF detection on the above mentioned evaluation dataset containing 175 coronary angiographies, obtained with a tolerance range of ±1 frames around the ground-truth EDF as detected based on the ECG signal. When compared to the results in Table 5, we note that the statistical measures are significantly better, which can be explained by the stricter exclusion criteria mentioned above.

Table 7: Statistical measures for EDF detection, with a tolerance range of ±1 frames around the ground-truth EDF as detected based on the ECG signal.

| Statistical measure | Performance |
|---|---|
| Precision | 99.3% |
| Recall | 99.3% |
| F1 score | 99.3% |

### 3.3. Cardiac phase detection runtime performance

For evaluating the execution time of the proposed workflow we considered a desktop computer with the following hardware configuration: Intel® Core™ i7-4820K CPU @ 3.70GHz, Nvidia TITAN X graphics card, 64GB RAM. Table 8 displays for the evaluation test set consisting of 175 angiographies the execution times for the online prediction phase. The average number of frames per angiography in this dataset was of 95.1 ± 14.9. Overall, the entire workflow for online cardiac phase prediction requires on average less than 5 seconds. The vesselness estimator has a longer execution time than the cardiac phase detector since it is applied for all frames, as opposed to the cardiac phase detector which is applied only for a subset of the frames.

Table 8: Execution times for the online cardiac phase prediction.

| Component | Execution time [s] |
|---|---|
| DL based vesselness estimator | 1.98 ± 0.48 |
| DL based cardiac phase detector | 0.75 ± 0.13 |
| Entire workflow, including pre- and post-processing | 4.9 ± 1.25 |

## 4. Discussion and Conclusions

Cardiac phase and end-diastolic frame detection are essential steps for the quantitative processing of coronary angiographies. Currently, the selection of the end-diastolic frame is performed either manually or automatically based on the simultaneously acquired ECG signal [5]. The ECG signal may not always be available, and the ECG-based cardiac phase detection has several drawbacks: the signal-to-noise ratio may be too low to accurately detect end-diastole or the signal may present artefacts [11], [12].

Herein we report, to the best of our knowledge, the first deep learning based workflow for purely image-based cardiac phase classification of angiographic frames validated on a large, real-world dataset. The angiographic images used in this study, except for the ones in the secondary evaluation dataset, were acquired consecutively, from patients examined in the catheterization lab, without any exclusion criteria related to patient state or history: the datasets were acquired from patients with stable CAD or acute CAD, previous myocardial infarction, previous PCI, previous CABG, coronary total occlusions, etc. Training data also included the natural variation in angiographic acquisitions of each view, including variations in image quality. We specifically avoided idealization of training and evaluation datasets, to ensure that the model is applicable in daily clinical practice with the accuracy reported herein. Both training and evaluation datasets were very large, consisting of a total of 45700 coronary angiographies acquired from 10150 patients. This ensures both model generalization and accurate prediction statistics.

One of the secondary design goals for the workflow was to ensure a short execution time, to not introduce any delays in routine clinical practice. We used a relatively small number of parameters for each DNN and obtained on average an execution time of less than five seconds for the entire workflow on a hardware configuration based on a Nvidia TITAN X graphics card.

We note that the cardiac phase detection performance on the main evaluation dataset was similar to that obtained on the subset of coronary angiographies for which the corresponding ECG signals were manually annotated. Thus, the results of the evaluation based on annotations performed by medical experts provided indirect validation of the workflow based on ground truth labels defined algorithmically.

Previous methods reported for automated cardiac phase detection on non-coronary medical images had similar or lower accuracy than the one reported herein. For example, the fully-automatic method for systole-diastole classification of real-time three-dimensional transesophageal echocardiography (RT-3D-TEE) data had a classification accuracy of 91.04% [14]. A deep learning model focusing on transthoracic echocardiogram (TTE) images reached a classification accuracy of 97.8% when using multiple images from each clip, and of 91.7% on single low-resolution images [15]. A method for automated cardiac resting phase detection in 2D cine MR images had a sensitivity, specificity, and PPV respectively at 73.2%, 90.6%, and 81.4% [20]. The results from these studies cannot be directly compared since different imaging modalities visualizing different cardiovascular structures were considered. We note that the workflow introduced herein can be directly applied to perform purely image based cardiac phase detection for any of these different imaging modalities or cardiovascular structures.

We further performed an error analysis for coronary angiographies extracted from the test datasets, for which the cardiac phase classification had poor performance. Several causes for suboptimal cardiac phase detection were identified, which are discussed in detail below.

We observed in several cases a phase shift between the predicted cardiac phase probabilities and the ground truth labels. In the majority of cases visual inspection of the angiographies and the predicted cardiac phase labels revealed that the onset of contraction and relaxation were predicted correctly, even in case of arrhythmias, but were not in sync with the start of systole and diastole as identified from the associated ECG signal. The desynchronization between ECG features and angiographic appearance may have been caused by cardiac conduction disorders, which in turn may be inherited [32], or may have an immunological basis [33]. Since we did not have access to patient characteristics or diagnoses for the main evaluation dataset, we could not verify this hypothesis. Other pathologies may have been present, which affect the ECG signals or the correct identification of ECG features: a heart attack occurring during the acquisition, changes to the heart's structure, stemming e.g. from cardiomyopathy, high blood pressure, diabetes, etc.

Another important cause for suboptimal cardiac phase prediction accuracy is the presence of heart tissue scarring, as caused by prior myocardial infarction(s). In these cases the ECG signal may not be affected, but the coronary artery contraction patterns may be abnormal.

Furthermore, the prediction performance was suboptimal in cases where very few arteries are visible on the angiographic images:
- Total occlusions: if one of the three main coronary arteries (LAD, LCx, RCA) is completely blocked, the number of coronary branches visible on the angiographies decreases significantly, making the task of identifying the contraction and relaxation more difficult
- Distal coronary artery inspection: in some angiographies only a limited part of the left or right coronary tree is visualized, especially when coronary stenoses are present only in the distal part of the epicardial arteries
- Stent inspection: excessive zooming may be performed to inspect stent deployment

The frame interval selected by the DL based vesselness model used a criterion based on a relative and not an absolute threshold. Hence the number of arteries visible in the angiography was not taken into account when selecting the frame interval of interest.

Finally, we noticed that in cases of panning the confidence and accuracy of the predictions decreases. Figure 8A displays the first frame selected by the vesselness estimator for an angiography with significant panning: two distal anatomical landmarks are considered, and their motion patterns throughout the frame interval selected by the vesselness estimator, which covers multiple heart cycles, is displayed. Figure 8B displays the last frame considered for this analysis: the anatomical landmarks have reached the end locations of the motion patterns displayed in Figure 8A. Finally, Figure 8C displays the cardiac phase probabilities output by the proposed workflow: the large difference between the probabilities and the target values of 0 and 1 indicates the low confidence in the predictions.

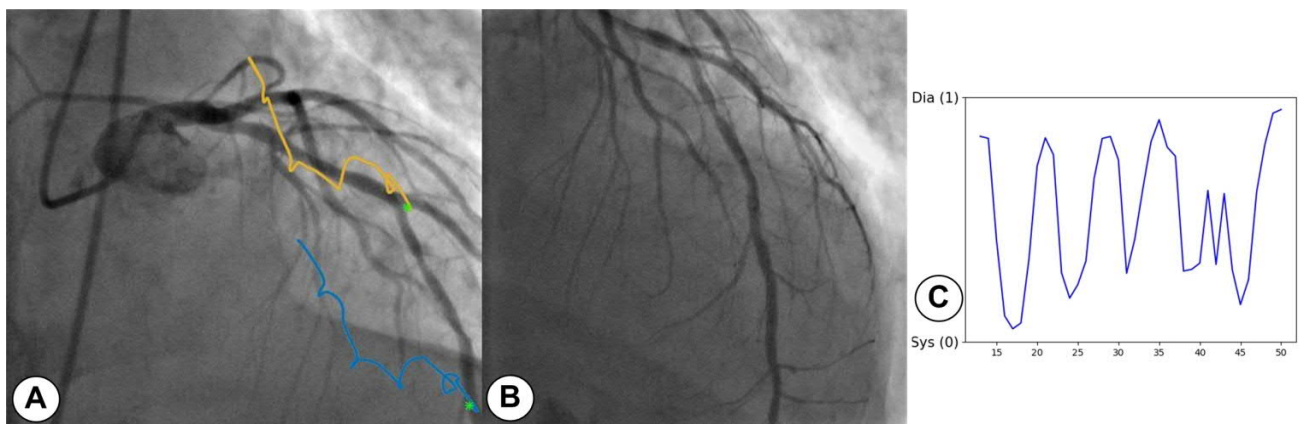

Figure 8: Coronary angiography displaying panning: (A) First frame selected by the vesselness estimator: two distal anatomical landmarks are considered, and their motion patterns throughout the frame interval selected by the vesselness estimator, which covers multiple heart cycles, is displayed; (B) Last frame selected by the vesselness estimator: the anatomical landmarks have reached the end locations of the motion patterns displayed in Figure 8A; (C) Cardiac phase probabilities for the frame interval selected by the vesselness estimator.

We performed several sub-group analyses, which indicated that the cardiac phase detection performance is largely independent from acquisition angles and the heart rate of the patient. Furthermore, detection accuracy was higher for the LCA than the RCA, suggesting that the detection accuracy increases with the number of arteries visible in the angiography.

As described in the previous section, the cardiac phase prediction results were significantly better for the secondary evaluation dataset consisting of the 175 coronary angiographies used in a recently published study focusing on FFR [31]. Since that dataset had strict exclusion criteria, several of the causes mentioned above for suboptimal performance, could be excluded. Specifically, all these angiographies had well visible coronary arteries, none of the acquisitions displayed panning, and coronary total occlusions were excluded. These aspects justify the ~1% increase in statistical performance, and the almost perfect EDF detection results.

We conclude that the proposed image based workflow, employing deep neural networks, demonstrated good performance, thus potentially obviating the need for manual frame selection and ECG acquisition, representing a relevant step towards automated CAD assessment[1].

**Appendix**

A deep learning based model is employed to determine the angiographic frame interval on which coronary arteries are well visible, i.e. filled with contrast agent. The classifier is a neural network with an architecture similar to U-Net [34], trained to classify pixels using the Adam optimizer [26] by minimizing a custom loss function derived from the Jaccard index:

$$L = 1 - \frac{\mu + \sum P_i T_i}{\mu + \sum P_i^2 + \sum T_i^2 - \sum P_i T_i},$$

where $P_i$ and $T_i$ are the predicted and the expected probability that the i-th pixel is part of a coronary artery, and $\mu = 0.1$ is a smoothing factor.

Training data are angiographic frames in which coronary vessels are manually annotated. Each artery is annotated as a set of centreline points, and to each point an approximate estimate of the local vessel radius is associated. Using these annotations, a binary mask is generated for each angiographic frame, in which pixels found in the neighbourhood of centreline points are set to 1; all other pixels are set to 0. The neighbourhood of the centreline points is defined as the locus of all pixels whose distance from the nearest centreline point is smaller or equal to the estimate of the local vessel radius.

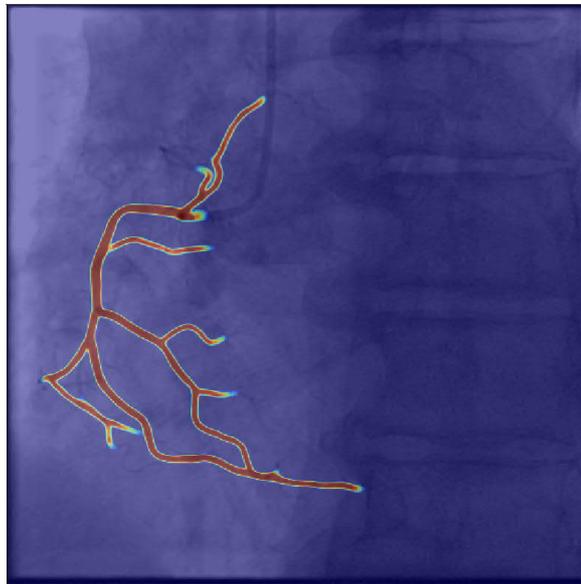

Figure 9: Overlay of the input frame and the predicted probability map.

Since the execution time was more important than the actual segmentation accuracy, a relatively small model with few parameters was designed. The model was trained on 1174 cases and validated on 293 cases. From each case, five consecutive frames were selected, resulting in 5870 training frames and 1465 validation frames. On a separate testing set consisting of frames extracted from 91 cases, the model achieved an average Dice score of 0.86. During inference, the model processes independently each angiographic frame and outputs a probability map with the same size as the input image, where the value of each pixel represents the probability that the associated pixel from the input image is part of a coronary artery, as shown in Figure 9. By summing all pixel-wise probabilities, an overall vesselness score of the frame is determined, i.e. the likelihood of the frame to display coronary vessels.

---

[1] This feature is based on research, and is not commercially available. Due to regulatory reasons its future availability cannot be guaranteed.

For each angiographic frame in the series, the vesselness score is determined. Frames having a vesselness score lower than two thirds of the maximum vesselness score in the series are discarded. From the remaining frames, the longest subsequence of consecutive frames is selected for further processing.